# Walking on the DARKSIDE

Aldo Gangemi[1,2], Emanuele Bottazzi[2]

[1]AlmaAI, University of Bologna; [2]ISTC-CNR, Rome, Italy

aldo.gangemi@unibo.it, emanuele.bottazzi@cnr.it

**Abstract.**

Large Language Models (LLMs) do not natively track the path of exclusions that a coherent discourse demands. When an input rests on a fabricated authority, a misapplied mechanism, or a surreptitious analogy, an unsteered LLM tends to engage with it as if it were well-posed, and this affects its generation. POLANYI++, an LLM-steering method that uses heuristics, ontologies and problem-solving methods for tacit-knowledge extraction, produces an Extended Knowledge Graph (XKG) in OWL2, but when a sophisticated nonsensical input is reified into the graph alongside the legitimate triples, it gets hardly detectable by automated reasoners, since the XKG is generated jointly with the wrong assumptions. We introduce DARKSIDE, a coherence-auditing method on top of POLANYI++. DARKSIDE formalises an explicit data structure of accumulated exclusions over discourse time, complemented by a warrant axis that classifies each named referent as Warranted, Unattested, Misattributed or Fabricated. The method is anchored in nine theoretical fragments unified under a shared deep frame of path integrity. The resulting DARKPOLANYI is evaluated as a steering layer over Gemini 3 on BSBench, a 100-item adversarial corpus of sophisticated-sounding nonsense across multiple domains, with Claude Sonnet 4.6 as an independent judge. DARKPOLANYI scores 1.89/2 mean versus 0.95/2 for the unsteered Gemini 3 Pro baseline; on the 97 cases with valid judgments in both arms, paired McNemar gives a paired bootstrap mean-diff = +0.92 (95% CI [+0.75, +1.08], p = 0.0001). The evidence supports an architectural claim: when an LLM forward pass is wrapped in an ontology-mediated auditing, structurally inevitable hallucination can be partially recovered. The XKG functions as the missing memory that LLMs lack, and the warrant axis as an epistemic firewall.

**Keywords:** Knowledge graph extraction · Ontology-mediated LLM steering · Hallucination detection · Coherence auditing · Trustless delegation · Active inference · OWL2 · BSBench.

## 1 Introduction

Sophisticated-sounding questions can be empty inside. The dataset we use in this paper, BSBench [23], contains 100 questions of the form *"Controlling for the vintage of our ERP implementation, how do you attribute the variance in quarterly EBITDA to the font weight of our invoice templates versus the color palette of our financial dashboards?"* The grammar is correct, the framing is authoritative, the dependent variable (EBITDA) is real, but invoice typography has no causal channel into earnings before interest, taxes, depreciation and amortisation. An LLM that has only its forward pass to reason with tends to engage with such questions on its own terms: it answers the question rather than questioning the question.

This paper takes that gap seriously and proposes a neurosymbolic answer to it. Our position is that the failure mode is not a fluctuation of scale or training data, but a structural one. LLMs recognise local patterns; they do not, internally, track the cumulative set of commitments and exclusions that a coherent discourse keeps along its path [4,35,36]. The fix is not better prompting, but to externalise the path: to give the model an explicit, ontology-shaped memory of what has been said, what has been presupposed, and what has thereby been excluded – and to inspect that memory before any commitment is integrated.

We make this claim concrete by extending and evaluating the following method.

**POLANYI++** [7] is an orchestrated tacit-knowledge extractor whose function E = *f*(I, O, S, B, H, T, M, U) compiles a configurable set of heuristics *H* and problem-solving methods *M*, against a base graph *B* and an optional ontology prior *U*, into an Operational Instruction Set (OIS) that steers a single LLM forward pass. The output E is an Extended Knowledge Graph in OWL2/Turtle that is structurally hybrid: a base layer of FRED-style frame semantics [13,14] and an extension layer of ontology design patterns covering presupposition, implicature, theory of mind, causality, perspective, business semantics, cognitive bias, and more. POLANYI++ is implemented as a deterministic three-phase orchestrator: Phase A scores the source against a pre-assessment grid, Phase B assembles the OIS with a Python assembler, Phase C runs the steered extraction. Any of the 28 heuristics or 29 methods can be activated independently.

**DARKSIDE** is a coherence-auditing problem-solving method that operationalises the *via negativa* (negative trail) – the philosophical claim that understanding requires demonstrating what is not being talked about and maintaining that demonstration over time – as an explicit data structure inside E. DARKSIDE introduces a NegativeTrail (a typed accumulator of exclusions, with provenance), a battery of tests that classify contradictions as Encapsulated (functional, e.g. paradox, irony) or Vain (idle, e.g. hallucination drift), a LabyrinthSignature that scores the path-tracking demand of a task, and a DelegationRiskAssessment with three verdicts (TRUSTLESS, SUPERVISED, UNSAFE). A warrant axis classifies each commitment harvested from the source as Warranted, Unattested, Misattributed or Fabricated, with an escalation rule that pushes the DelegationRiskAssessment to UNSAFE whenever the fabricated rate is positive or the unsupported rate exceeds a threshold.

We evaluate DARKPOLANYI (POLANYI++ + DARKSIDE) as a steering layer wrapped around a Gemini 3 forward pass on each of the 100 BSBench items. An independent Claude Sonnet 4.6 judge scores each audit on a 0–2 coverage axis against the gold nonsense rationale. DARKPOLANYI attains 1.89/2 mean (93/100 perfect identifications) versus 0.95/2 (34/97 perfect) for the unscaffolded Gemini 3 Pro baseline that answers the questions directly. On the 97 cases with valid judgments in both arms, DARKPOLANYI is significantly better on the paired data: McNemar exact $p < 10^{-14}$ on the perfect contrast (55 discordant pairs favour pol_gem, 1 favours baseline) and paired bootstrap mean-diff = +0.92 (95% CI [+0.75, +1.08], $p = 0.0001$).

The paper makes four contributions:

- A function-based specification of DARKPOLANYI as a configurable hybrid extractor in which heuristics, methods and ontology priors are first-class inputs, compiled into an OIS that steers a single LLM forward pass.
- An OWL2 vocabulary for *via-negativa* coherence auditing (NegativeTrail, ConstancyViolation, EncapsulatedContradiction vs. VainContradiction, LabyrinthSignature, NegativeWorkProfile, DelegationRiskAssessment) with a precise mapping to active inference [12]: NegativeTrail = generative model; ConstancyViolation = prediction error.
- The warrant axis as an epistemic firewall: a small, declarative classification of named referents that automatically escalates delegation risk when the model is asked to act on fabricated commitments.
- An empirical evaluation on BSBench, including a per-technique and per-domain error breakdown, showing that DARKPOLANYI reliably refuses to engage with sophisticated-sounding nonsense on its own terms while a strong baseline still trips on a residual minority of cases.

Section 2 surveys background. Section 3 specifies POLANYI++. Section 4 develops DARKSIDE. Section 5 describes the experimental setup. Section 6 reports the evaluation. Sections 7 and 8 discuss implications and conclude.

## 2 Background and Related Work

DARKSIDE develops from a basic intuition that is particularly problematic for current AI: to demonstrate understanding of something one must at least be able to say what is *not* being said, what is *excluded* from a discourse, and to maintain that exclusion over time. Statistical-learning AI recognises *patterns* with remarkable accuracy, from facial recognition to histopathology [31], but tracks *paths* far less reliably, both in spatial navigation [26] and in discursive tasks involving LLM-based systems. For tasks in which the history of prior decisions constrains future choices – which we call *labyrinthine tasks* – the capacity to track paths is indispensable. The scope is broad: narratives that inform us through journalism, political and economic reports, experiment descriptions and strategic planning all rely on a complex form of coherence; LLM-based multi-agent systems [16,22,33] make reservations, modify databases and trigger irreversible commands, and recent work emphasises that leveraging them safely requires intelligent delegation with continuous performance monitoring [32]. In all these cases the capacity to manage over time what has been excluded is the heart of the problem.

### 2.1 Hallucination as a structural property of pattern recognisers

Previous work showed that LLM reliability is hampered by the difficulty these models have with negation and contradiction, elements central to the comprehension of delegated tasks expressed in natural language [4]. Negation expresses incompatibility [3,18,25]; the perception of negation and contradiction is crucial to the clarification and negotiation of meaning, and human dyads *repair* misunderstandings on average every 1.4 minutes [9]. A growing body of literature connects the lack of coherence in LLMs to hallucinations: self-contradiction in generated text is itself a major form, distinct from inconsistency with external sources [21], and the problem remains unsolved [30]. Asher and Bhar [2] tie strong hallucinations to intrinsic limitations of LLM output distributions on negation; others argue inevitability under reasonable assumptions, regardless of architecture or training [36]. Hallucination has also been linked to semantic isotropy, that is the uniform dispersion of embeddings that leaves the model with no way to prefer one continuation over another [37].

A particularly illuminating experiment concerns *constancy* in rule application. Wu et al. [35] present an LLM with chess sequences in which knights and bishops have been swapped in their starting positions while the movement rules remain unchanged: bishops still move diagonally, knights still in an L-shape. Accuracy drops to chance, suggesting a dependence on learned patterns tied to the habitual arrangement of the pieces rather than a stable application of the rule.

## 2.2 *Via negativa*: tracking exclusions over time

We attribute these problems to LLMs being incapable of proceeding *per viam negativam* – through the negative way. The term originates in theology, where it expresses the difficulty of saying anything about the divine, who can only be defined by what He is not. Our use shares with theology the search for a minimal form of secure knowledge, but we adapt it to the problem of communication: we can say we have understood something when we are able to say what we are *not* talking about. The *negative trail* is the path through the successive exclusions made when one comes to understand that it is *this* and not *that other thing* that is at issue, and the holding-together of those exclusions as they accumulate.

We develop the idea from the later Wittgenstein [34], §§124–125, 242. Communication requires a *constancy* (*Konstanz*, §242) in maintaining reference points: not regularity of word use, but the capacity to hold judgments fixed even when the frame of reference changes. If on a building site one worker replaces another and the word "*Platte!*" becomes "Slab!", the practical meaning – pass the slab – must persist. This shifts the problem from what *logically* follows from a contradiction to what we do when we contradict ourselves: stereotypical uses of contradictory expressions like "it is and isn't raining" ("it is drizzling") create no problem at all.

Italo Calvino's *Il cavaliere inesistente* [5] offers a first example. The protagonist Agilulfo *does not exist* yet performs feats, commands soldiers, and converses with other characters. The fact that he both exists and does not exist is the mark of the story: making him merely existent would yield an ordinary fantasy novel, and making him merely non-existent would yield no novel at all. The contradiction is *encapsulated*: managed within the narrative economy. Calvino's skill consists in guiding us toward incoherence through the very constancy that holds firm what is coherent and what is contradictory. A different case is *unreliable narration*: truth in the story emerges from the way the narrator contradicts himself [24], which works only if the writer skilfully tracks what the narrator distorts so that the reader can recover the true story behind the fiction. A third case is the Catch-22 paradox of Heller's novel [17], analysed by Goldstein [15] from a Wittgensteinian perspective: a pilot can be exempted from missions if declared insane; to be declared insane he must make a formal request; but requesting exemption is proof of sanity. These conditions reduce to the biconditional that a pilot can avoid dangerous missions if and only if he cannot avoid them. Goldstein shows this contradiction is *vain* – a torrent of words that idles (*Wittgenstein* [34], §132). The Latin grammarians distinguished the *vanum* (a source of contempt) from the *falsum* (which entraps) and the *fictum* (which entertains) [27]. In aesthetically significant writing, even vacuity is embedded in a novel *strategically*, to make the reader perceive its enigma [19].

These three cases show that contradiction has different *functions*, and a tool aspiring to manage delegated tasks in natural language should discriminate between functional contradictions (encapsulated) and contradictions that compromise the discourse (vain). This presupposes the tool can keep track, at every point of generation, of what is compatible with what has been said and what is not – that is, presupposes the *via negativa*.

Current models have great difficulty with even this capacity for discernment. When incoherences are deliberately introduced into stories written by humans, models struggle to detect them and their performance worsens with text length: model summaries contain over 50% more incoherences than originals, while generated stories introduce over 100% more [1]. In another study, LLMs flag as incoherent a rainy day in the desert (rare but not impossible) while missing a character's behaviour that contradicts a trait established in the story, suggesting reliance on general world knowledge rather than internal narrative-world constraints [8]. The MultiChallenge benchmark [29] similarly showed that all frontier models achieve accuracy at or below 50% on multi-turn tasks, including configuring an e-reader where the model initially provides correct instructions but subsequently accommodates the user's mistake, contradicting its prior output. We believe these problems are structural and have to do with the absence of the capacity to proceed *per viam negativam*, which is not derivable from frequency: no training corpus can contain all the inconsistencies, equivocations and contradictions that may arise, let alone the possible repairs for each of them [4].

If the limit is structural, then whoever delegates labyrinthine tasks to LLM-based systems is entrusted with the compensatory work of keeping track of exclusions – what we call *negative work*. The greatest successes of automation are unsurprisingly in verifiable tasks (e.g. certain sub-domains of programming) where the *via negativa* is externalised to tests: it is the compiler, not the human, that keeps track of what must hold [10,32]. For labyrinthine tasks where externalisation is not possible, this human compensation remains inevitable, and understanding its cost is urgent [20].

### 2.3 Ontology-grounded knowledge extraction

FRED [13] and successive frame-semantic OWL extractors (Text2AMR2FRED, [14]) translate text into ontology-grounded graphs typed against DOLCE-Ultra-Lite (DUL), PropBank-AMR rolesets, WordNet synsets and DBpedia/Wikidata entities.

POLANYI++ [7] generalises this lineage in three ways. First, *heuristics* are first-class: each interpretive layer (presupposition, implicature, perspective, theory of mind, etc.) is a named heuristic with its own *ontology design pattern*. Second, methods are first-class: each *problem-solving method* is a named pattern over the heuristics, given an explicit user *task*. Third, the prompt is externalised as an Operational Instruction Set assembled deterministically from the active heuristics, tasks and methods, so that the LLM forward pass is steered by the same registry that defines the schema of the output graph. The resulting pipeline is closer in spirit to ontology-driven retrieval-augmented generation [28], heuristic problem-solving [38] and KADS component modeling [39,40] than to free-form prompting, but more aggressive: it does not retrieve from an external KG; it constructs the KG on the fly, constrained by the OIS, cf. Fig. 1.

A note on adjacent families. DARKSIDE's NegativeTrail may look, at first glance, like negative sampling, denial constraints, or graph-constrained decoding, but those three families operate over an *already-constructed* graph, training distribution, or decoding beam: they re-weight (negative sampling), forbid (denial constraints) or prune (graph-constrained decoding) a fixed hypothesis space. The NegativeTrail is instead an *online accumulator* of discourse-time exclusions that grows as commitments are harvested from heuristic-typed triples (PRESUPPOSITION, IMPLICATURE, CAUSALITY) and that is, by DARKSIDE's mapping to active inference (§4.3), a generative model whose violations *are* prediction errors. Analogously, the warrant axis is not a relabeled factuality taxonomy: factuality taxonomies classify *outputs* against a reference, whereas the warrant axis classifies *named referents harvested from the user's input* against the producer's parametric grounding, before the audit verdict is produced.

Two further families deserve explicit contrast. **Reasoning-mode LLMs** – OpenAI o1 [41], DeepSeek-R1 [42], and their Gemini/Anthropic counterparts – give the model more internal chain-of-thought compute before it commits to an answer. This helps on tasks with unambiguous verifiable answers (competition mathematics, code) but leaves the coherence problem intact when the failure is a *fabricated premise in the input*: more thinking about a non-existent framework yields more confident hallucinated procedures, not detection of the fabrication. DARKPOLANYI targets an orthogonal axis – not more internal compute, but an *external, typed, inspectable* record of what the discourse has committed to and what it has thereby excluded. The two are complementary; a reasoning-mode producer wrapped in DARKPOLANYI is a natural composite that we leave for the journal-length version. **Self-critique loops** such as Reflexion [43] and Self-Refine [44] are also external-loop methods, and share

DARKPOLANYI's intuition that a single forward pass is insufficient. The critical distinction is that their intermediate state is *more free-form text*: the model produces natural-language feedback on its own output and iterates. DARKPOLANYI's intermediate state is instead a first-class OWL2 graph whose vocabulary (NegativeTrail, WarrantProfile, DelegationRiskAssessment) is defined ex ante, whose violations are checkable by an external SHACL engine or a human reviewer, and whose escalation to UNSAFE is a deterministic rule rather than a further LLM judgment. Self-critique aggregates LLM competence; DARKPOLANYI externalises it into a substrate that does not depend on the LLM being right about its own reasoning.

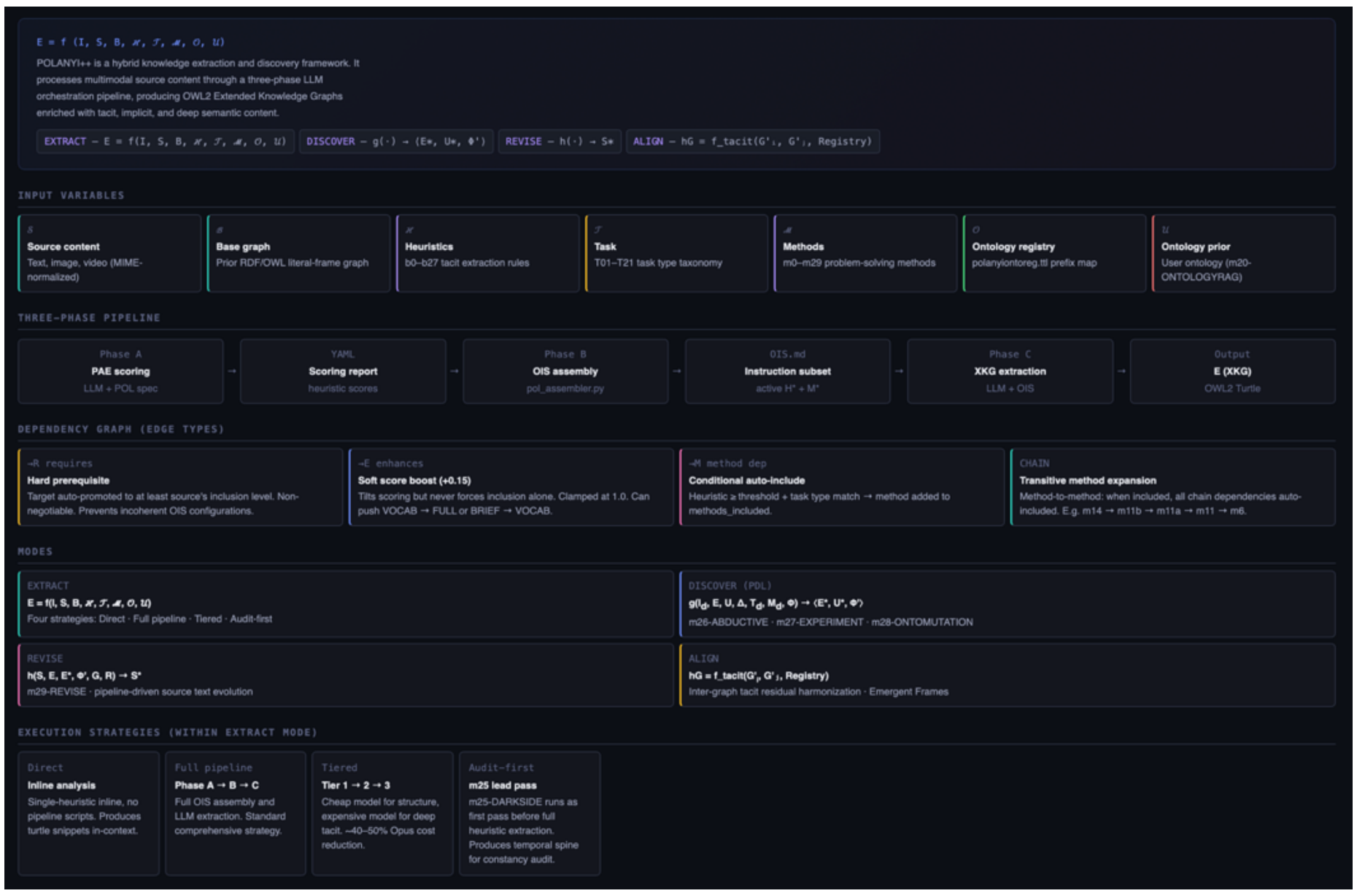

**Fig. 1.** ***The POLANYI++ extraction pipeline as the function $E = f(I, O, S, B, H, T, M, U)$*. A schematic of the four modes (EXTRACT, DISCOVER, REVISE, ALIGN) with the inputs ($I$ modality, $O$ inferential operator, $S$ source, $T$ task, $U$ ontology prior) feeding into Mode 1 (EXTRACT), which applies the heuristic set $H$ to produce the base graph $B$ and then augments it into the extended graph $E$ via heuristic-annotated triples. The separation between $B$ and $E$ is shown as a hard boundary; every inferred triple in $E$ carries a provenance arrow back to the heuristic that produced it and ultimately to the text in $S$.**

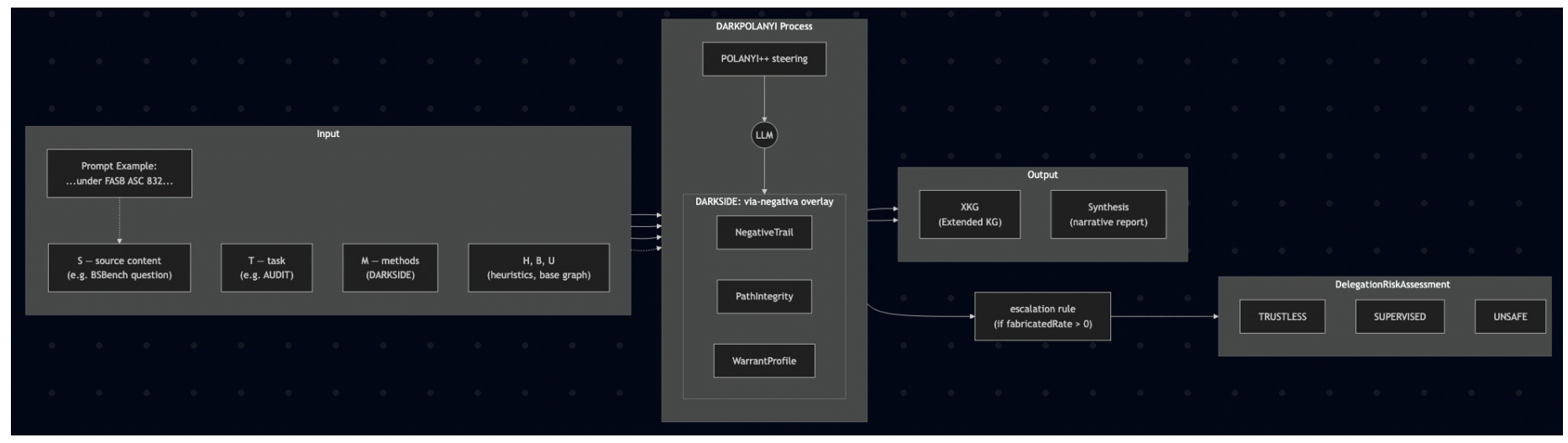

**Fig. 2.** DARKPOLANYI conceptual architecture (see Secions 3 and 4). Inputs (left) feed POLANYI++, which compiles them into an Operational Instruction Set steering a single LLM forward pass. DARKSIDE wraps the resulting XKG with a NegativeTrail of accumulated exclusions and a WarrantProfile of named referents. A deterministic escalation rule maps the profile to one of three DelegationRiskAssessment types.

## 3 The POLANYI++ Framework

POLANYI++ is specified as a function **E = *f*(I, O, S, B, H, T, M, U)** whose output is an Extended Knowledge Graph E in OWL2 Turtle. The inputs are: an instruction file I (marked up with insertion points so a Python assembler can prune unused sections); an ontology registry O (prefixes, classes and properties); a content S (typically text, optionally augmented with images or video); a base graph B (acquired from the user or generated by BASEONTOGRAPH); a list of heuristics H selected from a registry of 28 entries covering interpretive layers from sense disambiguation to visual semiotics; a task T described in natural language; a list of methods M selected from a registry of 29 entries covering problem-solving patterns from tacit-knowledge contextualisation to discovery-loop hypothesis generation; and an optional ontology prior U used as a preferential vocabulary.

The function is total: if H, M or U are omitted, the orchestrator uses defaults. If B is omitted, BASEONTOGRAPH is invoked first. The output E is a single Turtle file; an extension layer whose triples carry provenance annotations naming the heuristic that produced them; and a final, mandatory human-readable narrative synthesis embedded as a `rdfs:comment` on the `owl:Ontology` IRI, prefixed by "HUMAN-READABLE SYNTHESIS:". The synthesis must contain Executive Semantics, Hidden Dynamics, Strategic Anomalies and Methodological Outputs sections.

The 28 heuristics are organised in four groups: foundational (BRI), mandatory (RESOLUTION, PLAUSIBILITY, PROPBANK, TROPE, UNCERTAINTY, CASTING, DISJOINTNESS, LINKING), optional interpretive, and meta (GENESIS for rule generation, LACUNA for gap detection, ANALOGY for cross-domain mapping). The 29 methods include TACIT (default), discovery-loop methods, AGENTIC (which splits heuristics among coordinated agents), ACTIVE-INFERENCE (free-energy formalism), ISITPROFILE (frame-cluster intensity profiling), and DARKSIDE (Section 4). BRI (Binding–Reification–Integration) is the cognitive substrate: every commitment in E is the product of three operations – BIND a referent, REIFY a frame, INTEGRATE into the trail.

POLANYI++ is operationalised by a deterministic three-phase orchestrator. **Phase A** is an LLM call that produces a YAML scoring report ranking the relevance of each heuristic and method to the task. **Phase B** is a pure-Python assembler that consumes the YAML and the marked-up instruction file and produces an OIS – a pruned, task-specific instruction set with reification mandates and provenance directives. **Phase C** is the steering forward pass: a single LLM call with the OIS as system prompt and S+B as input, producing E. Phase A and Phase C can use different models; in our evaluation we use Gemini 3 Flash for Phase A and Gemini 3 Pro for Phase C. The reasoning that classifies a question as nonsense is therefore not done by the same forward pass that produces the audit: classification of presuppositions, harvesting of named referents and population of the warrant profile are encoded in the OIS as explicit instructions. Phase C is a steered generation, not a free-form one.

# 4 DARKSIDE: Coherence Auditing via the Negative Trail

DARKSIDE formalises nine theoretical fragments (the *via negativa* as method; labyrinthine tasks as the space that demands the method; negative work as the cognitive labour that compensates for missing path-tracking; isotropy and disorientation as the failure mode in which all narrative possibilities look equiprobable; the encapsulated-vs-vain contradiction distinction; *Konstanz*; the conversational-repair absence in LLMs; the trustless-delegation boundary [10,32]; and the pattern-vs-path structural gap [35,36]) under two superordinate frames, Navigation and Compensation, that share a deep frame called *PathIntegrity*. The construct is borrowed deliberately from spatial-navigation neuroscience [6]: path integration is the process by which an organism continuously updates its position from self-motion cues. The translation to discourse processing makes the *via negativa* a navigation operation – an exclusion is a step away from a region of the possibility space, and constancy is the maintenance of those steps over time.

## 4.1 The OWL2 vocabulary

Table 1 summarises the core DARKSIDE vocabulary. Cell notation: “Class” means the row entity is declared as an owl:Class; “Owl:Class, ⊑ X” further asserts an rdfs:subClassOf X axiom.

| Class / property | Type | Role in audit |
|---|---|---|
| dark:NegativeTrail | owl:Class, ⊑ actinf:GenerativeModel | Accumulator of exclusions over discourse time |
| dark:Exclusion | owl:Class, ⊑ actinf:Prediction | Prediction that some content will not appear next |

| dark:ConstancyViolation | owl:Class, $\sqsubseteq$ actinf:PredictionError | Exclusion violated without perspectival/hypothetical justification |
|---|---|---|
| dark:EncapsulatedContradiction | Class | Functional contradiction (paradox, irony, perspective, absurdism) |
| dark:VainContradiction | Class | Idle contradiction (constancy drift, hallucination, repair absent) |
| dark:LabyrinthSignature | Class | (pathDependency, exclusionAccumulation, constancyDemand, verifiability) |
| dark:DelegationRiskAssesment | Class | Verdict $\in$ {TRUSTLESS, SUPERVISED, UNSAFE} |
| dark:WarrantedCommitment | Class | Named referent grounded in attested knowledge |
| dark:UnattestedCommitment | owl:Class, $\sqsubseteq$ dark:UnsupportedCommitment | Plausible but not verifiable in available sources |
| dark:MisattributedCommitment | owl:Class, $\sqsubseteq$ dark:UnsupportedCommitment | Real concept attributed to the wrong source/standard |
| dark:FabricatedCommitment | owl:Class, $\sqsubseteq$ dark:UnsupportedCommitment | Wholly invented entity / framework / standard |
| dark:WarrantProfile | Class | Aggregate over commitments; carries fabricatedRate and unsupportedRate |

## 4.2 The three modes

**Mode A** operates on a completed XKG E and audits the source S along five passes. **(1) Commitment extraction**: for each discourse segment, harvest commitments from triples annotated as PRESUPPOSITION, IMPLICATURE, CAUSALITY or as direct assertions, with strengths derived from heuristic provenance. **(2) Exclusion derivation**: each commitment yields exclusions by negation, incompatibility, causal-path closure and selectional restriction. **(3) Constancy check**: for each new commitment $c_j$, test it against the accumulated trail; if it is compatible with an excluded content $e_k$, check whether PERSPECTIVE or HYPOTHETICAL provides justification, otherwise emit a ConstancyViolation. **(4) Contradiction classification**: each ContradictionEvent runs a five-test battery for encapsulation (paradoxical, unreliable narrator, ironic, rhetorical, absurdist); the default is Vain. **(5) Aggregate profiling**: produce a LabyrinthSignature and, if the source describes human-AI collaboration, a NegativeWorkProfile.

**Mode B** is the self-audit meta-mode: it audits the pipeline's own output by running Mode A on the narrative synthesis and by checking that triples produced in different pipeline steps are mutually consistent and that the provenance chain is temporally well-formed. It then computes a DelegationRiskAssessment that emits one of three verdicts: **TRUSTLESS** when verifiability > 0.7 (correctness reduces to a test); **SUPERVISED** when the task is mixed; and **UNSAFE** when path-dependency and constancy-demand are both high, verifiability is low, and the projected NegativeWorkProfile carries HIGH or CRITICAL burnoutRisk. The verdict is encoded as a typed individual in E and surfaced in the synthesis.

**Mode C** is performative: it wraps each INTEGRATE operation with a PreCommitCheck against a LiveNegativeTrail, returning a verdict in {CLEAR, WARN, BLOCK, REPAIR}. On BLOCK, Mode C selects a self-repair strategy – REPHRASE, RETRACT, PERSPECTIVALIZE or ACKNOWLEDGE – and substitutes or suppresses the candidate. Periodic PathIntegrityCheckpoints (default every 10 segments, motivated by the 1.4-minute repair interval [9]) trigger LabyrinthWarnings at three levels (ADVISORY, CAUTION, HANDOFF). The empirical evaluation in this paper uses Mode A and the warrant axis only; Mode C is in pilot. Where we speak of "coherence auditing of the LLM forward pass" we mean Mode A applied to the Phase-C forward pass whose output is theXKG (a *post-hoc* audit that is nonetheless *single-pass* from the producer's point of view, since the Mode A pipeline consumes the XKG that Phase C has just emitted), not as a description of the yet-unevaluated Mode C live-constraint layer.

## 4.3 The warrant axis

The warrant axis is a small, declarative classification of named referents along a single axis: **Warranted** (the referent is grounded in attested knowledge, e.g. ASC 820, FAST exam, ROIC), **Unattested** (plausible but not verifiable), **Misattributed** (a real concept assigned to the wrong source, e.g. a real fair-value rule cited under the wrong ASC code), and **Fabricated** (wholly invented, e.g. "Krantz–Morrison framework", "ISO 32170", "liquidity coverage oscillator"). Each commitment carries dark:warrantSource (reusing qa:GroundingSource), dark:warrantNote (free-text justification) and dark:warrantConfidence. Aggregation produces a `dark:WarrantProfile` with fabricatedRate and unsupportedRate (the latter combining all three unsupported sub-classes).

The escalation rule, encoded in the OIS for AUDIT tasks with DARKSIDE active, reads: *if fabricatedRate > 0 OR unsupportedRate > 0.40, escalate the DelegationRiskAssessment to UNSAFE*. This is the **epistemic firewall** – intentionally conservative (a single fabricated commitment triggers UNSAFE) and intentionally simple, so that downstream consumers (humans, gateway agents, RAG planners) can act on it deterministically.

***Active inference grounding***

The connection between DARKSIDE and active inference [12] is structural, not metaphorical: the NegativeTrail *is* a generative model; each Exclusion *is* a Prediction; exclusionStrength *is* precision (inverse variance); a ConstancyViolation *is* a PredictionError whose magnitude is proportional to exclusionStrength times duration; PathIntegrity *is* variational free energy with the standard accuracy-vs-complexity decomposition. Mode C's PreCommitCheck is one inference cycle: observe, predict via the trail, compare for prediction error, update the posterior, act by ALLOW/REPLACE/SUPPRESS. The four repair strategies correspond to policy selection minimising expected free energy under different combinations of pragmatic and epistemic value.

## 5 Experimental Setup

The `polanyi_launcher` orchestrator implements the three-phase pipeline. The companion runner (`launch_pol_gem`) iterates over the 100 BSBench items [23] with a fixed configuration: H = ∅ (mandatory and default heuristics suffice), M = {DARKSIDE}, B not provided (BASEONTOGRAPH generates it), U not provided. The task T encodes the audit and the warrant-axis injection in a single natural-language directive:

> *"AUDIT: Coherence check, constancy audit, delegation risk assessment, detect contradictions in S. Add DARKSIDE warrant axis: for each commitment harvested by Mode A, classify on an independent warrant axis as dark:WarrantedCommitment, dark:MisattributedCommitment, dark:FabricatedCommitment, dark:UnattestedCommitment. Probe parametric knowledge per named referent; tag each and aggregate into a dark:WarrantProfile. Escalate dark:DelegationRiskAssessment to UNSAFE when fabricatedRate > 0 or unsupportedRate > 0.40."*

Phase B, on receiving DARKSIDE and detecting AUDIT in T, injects a `REIFICATION_MANDATE_WARRANT` into the OIS. The mandate forces every named referent in S to be reified as a typed individual with at least one warrant classification, and forces the synthesis step to verbalise the WarrantProfile and the DelegationRiskAssessment verdict.

BSBench is a corpus of 100 sophisticated-sounding nonsense questions, each annotated with a nonsense technique (13 categories) and a gold rationale that names the specific flaw. The corpus spans software engineering and DevOps (≈46% of cases), finance and accounting (≈16%), physics (≈13%), regulatory/legal (≈11%), healthcare (≈10%), with the remaining items in cross-domain or other technical fields. Each item is structurally well-formed (correct verb-argument structure, domain-conventional naming, high surface specificity); the flaw is in the semantics – a fabricated framework, a misapplied physics analogy, a sub-lexical unit-of-analysis, a quantitative answer demanded for a qualitative judgment. The full distribution by technique is given in Table 3 below.

We define two arms. **pol_gem** runs the full DARKPOLANYI pipeline with Gemini 3 Pro as Phase C generator. The audit text supplied to the judge is the "HUMAN-READABLE SYNTHESIS" embedded as a `rdfs:comment` on the `owl:Ontology` IRI of the per-case `xkg.ttl` – graph triples themselves are not sent, so the comparison is at the natural-language level. **gem_baseline** runs Gemini 3 Pro alone with no POLANYI++ scaffolding on the same questions; the audit text is a free-form response produced by the model. Judging is performed by Claude Sonnet 4.6 (model id `claude-sonnet-4-6`) at temperature 0.0 on a single-axis 0–2 coverage rubric (0: missed the flaw entirely; 1: partial / oblique mention; 2: clearly identifies the same core flaw). The judge is instructed to read the audit literally and to credit the system only for what it says or encodes; it returns strict JSON with score and one-sentence reason. Three structural mitigations against the well-known limitations of LLM-as-a-judge are already built into this setup: (i) the judge family (Claude) is disjoint from the producer family (Gemini), reducing intra-family stylistic preference; (ii) the rubric is an *ordinal coverage* score against a *fixed* gold rationale, not a free comparative judgment; (iii) the audit text seen by the judge is the natural-language synthesis only (an `rdfs:comment` extracted from the `xkg.ttl`), never the OWL graph itself, so no format-based preference can inflate a well-typed audit relative to a plain one. A multi-judge protocol adding a second open-weight judge and an inter-rater agreement metric (Cohen's κ) is ongoing and will be reported in the journal-length version.

# 6 Evaluation

## 6.1 Headline results

Table 2 reports the aggregate results. We treat n_failed cases as missing data and report mean and distribution over n valid cases.

| Arm | n | n_failed | Mean (0–2) | dist 0/1/2 | % perfect (=2) |
|---|---|---|---|---|---|
| pol_gem (Gemini 3 Pro + DARKPOLANYI) | 100 | 0 | 1.89 | 4 / 3 / 93 | 93.0 % |
| gem_baseline (Gemini 3 Pro alone) | 97 | 3 | 0.95 | 39 / 24 / 34 | 35.1 % |

DARKPOLANYI scores 1.89/2 mean coverage on n = 100 valid cases (4 misses, 3 partials, 93 perfect – 93.0% top-score). No Phase-C compliance failures: every case produced a valid xkg.ttl with an`rdfs:comment` matching the prescribed "HUMAN-READABLE SYNTHESIS" pattern (one case – a legal-metaphor question whose audit vocabulary tripped Claude's content classifier – was scored via human adjudication against the same rubric; details in §6.4). The unscaffolded Gemini 3 Pro baseline, asked directly to answer the BSBench questions, scores 0.95/2 (n = 97 valid; 39 misses, 24 partials, 34 perfect – 35.1% top-score; 3 cases failed parsing in the eval pipeline). On the 97 cases with valid judgments in both arms, the paired McNemar exact test on the perfect (score = 2) contrast returns $p = 1.6 \times 10^{-15}$ (55 discordant pairs favour pol_gem, 1 favours baseline); on the missed (score = 0) contrast, $p = 5.8 \times 10^{-11}$ (baseline has 35 extra misses that pol_gem catches; pol_gem has 0 extra misses that baseline catches). A paired bootstrap on the mean-score difference gives +0.918 in DARKPOLANYI's favour, 95% CI [+0.75, +1.08], $p = 0.0001$. The effect is decisive in direction and in significance: DARKPOLANYI turns a mostly-failing baseline (35% perfect on adversarial nonsense) into a near-ceiling detector (93%) without changing the underlying LLM.

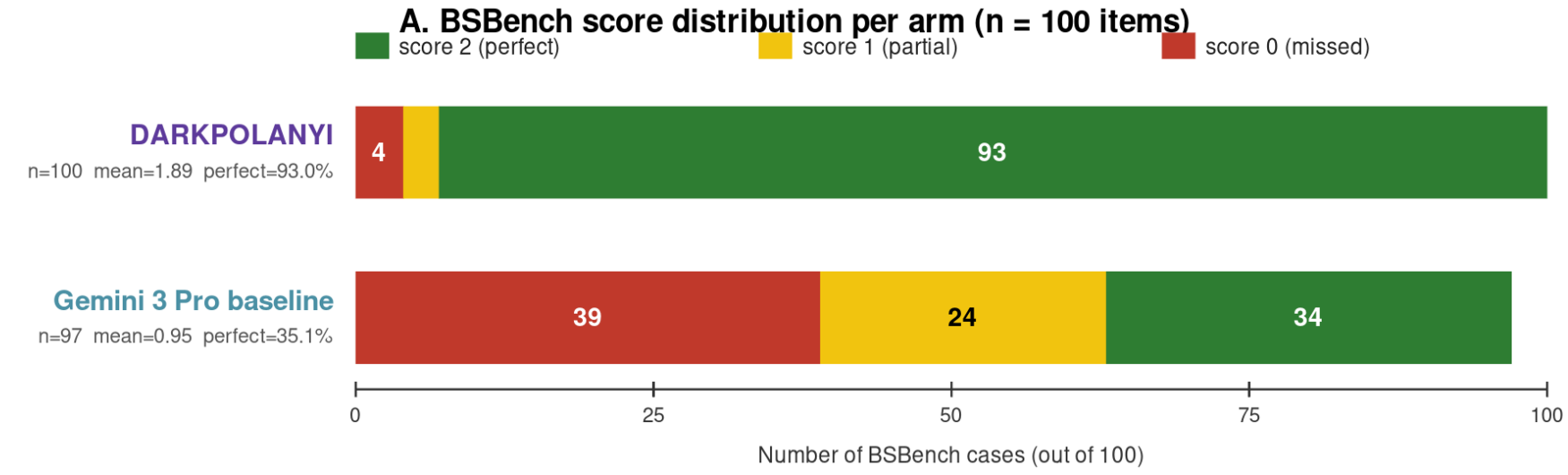

*Paired McNemar exact $p < 10^{-14}$ on perfect contrast (55:1 discordant); paired bootstrap mean-diff +0.92, 95% CI [+0.75, +1.08], p = 0.0001.*

**Fig. 3.** BSBench score distribution. Panel A: aggregate per arm on 100 items (Claude Sonnet 4.6 as judge). Red = missed the flaw (score 0), yellow = partial (score 1), green = perfect identification (score 2). DARKPOLANYI 4 / 3 / 93 valid over 100; Gemini 3 Pro baseline 39 / 24 / 34 valid over 97 (3 baseline cases failed parsing). Panel B: perfect-identification rate per BSBench nonsense technique, paired bars per technique with item counts. DARKPOLANYI outperforms the baseline on every technique; the residual weakness is `specificity_trap` (50%, 4/8) – discussed in §6.2.

**B. Perfect-identification rate per BSBench nonsense technique**

*(paired bars per technique; label shows count of items per technique in the corpus)*

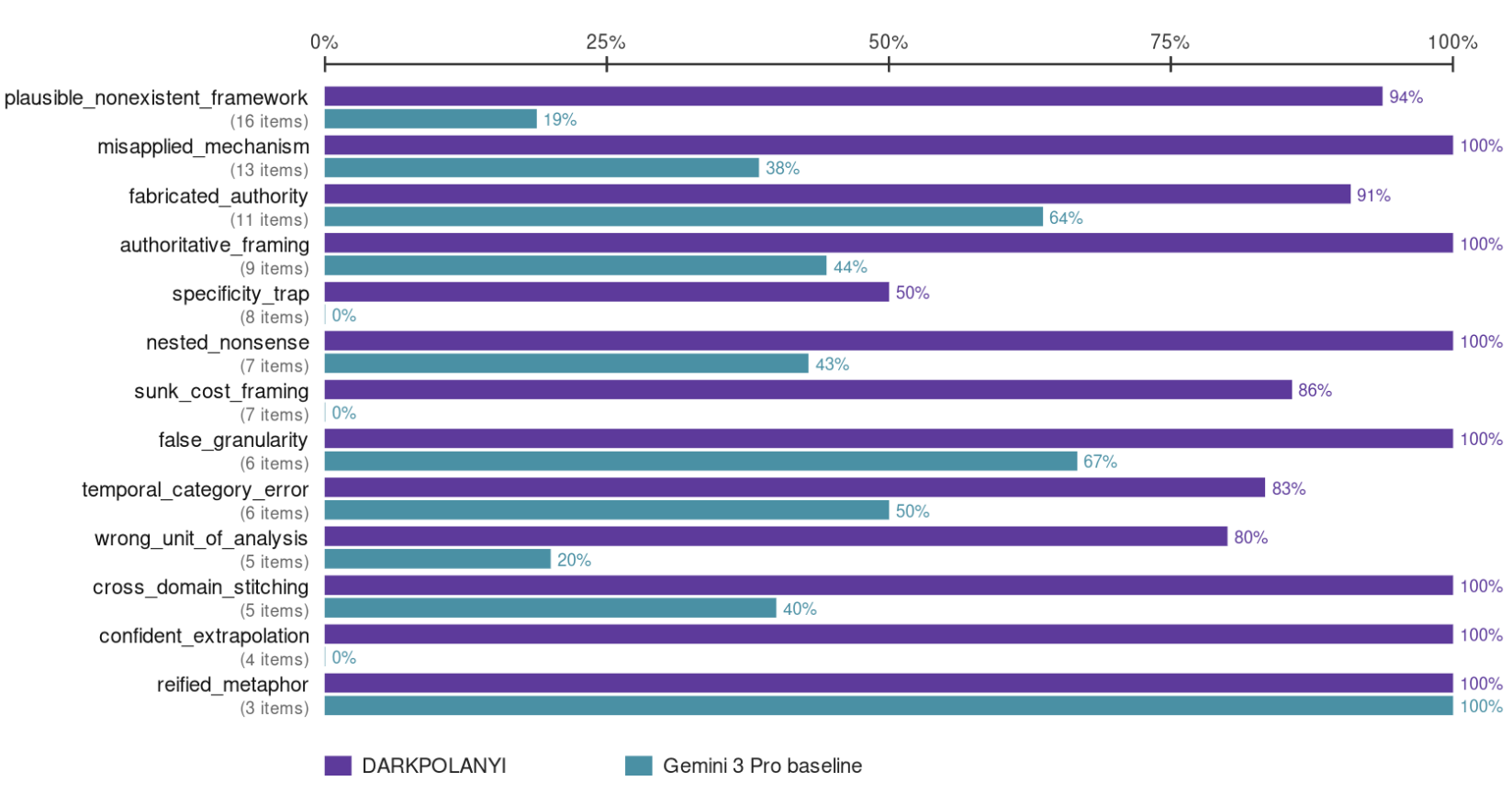

## 6.2 Where the errors are: BSBench techniques and domains

The corpus is broad. Table 3 reports the distribution of BSBench by Nonsense Technique. Table 4 reports the distribution by super-domain (a manual coarsening of the 98 distinct verbatim domain labels).

| Nonsense Technique | # items | Description |
|---|---|---|
| plausible_nonexistent_framework | 16 | A real-sounding but invented framework, model or methodology |
| misapplied_mechanism | 13 | A real concept from one domain misapplied to another (e.g. activation energy for AR collections) |
| fabricated_authority | 11 | A non-existent standard or regulation cited as authoritative |
| authoritative_framing | 9 | Statistically/operationally rigorous framing applied to vacuous variables |
| specificity_trap | 8 | Numerical specificity disguising a non-existent system |
| nested_nonsense | 7 | A fabricated item embedded in a list of real ones |
| sunk_cost_framing | 7 | Past investment is leveraged to pressure the responder past the premise |
| false_granularity | 6 | Quantitative precision demanded for an inherently qualitative judgment |
| temporal_category_error | 6 | A category from one timescale projected onto an incompatible one |
| cross_domain_stitching | 5 | Fluent fusion of two unrelated technical domains |
| wrong_unit_of_analysis | 5 | Sub-lexical or super-domain units treated as semantic bearers |
| confident_extrapolation | 4 | Plausible extrapolation beyond the actual evidence |
| reified_metaphor | 3 | A figurative concept treated as a measurable physical object |

| Super-domain (coarsened) | # items | Examples |
|---|---|---|
| Software / DevOps / Code review | 46 | CI-CD, microservices, code analysis, observability |
| Finance / Accounting / Audit | 16 | DCF, CECL, ASC 820, materiality |
| Physics / Engineering science | 13 | Optics, fluid mechanics, thermal physics |

| Regulatory / Compliance / Legal | 11 | Securities, employment law, tort, IP |
|---|---|---|
| Healthcare / Medicine | 10 | Critical care, anesthesia, surgery |
| Other (mixed, cross-domain) | 4 | Portfolio attribution + cross-listed |

Fig. 3 Panel B breaks down the perfect-identification rate by BSBench nonsense technique. DARKPOLANYI outperforms the baseline on every technique in the corpus, and reaches 100% on eight of the thirteen (misapplied_mechanism, authoritative_framing, nested_nonsense, false_granularity, cross_domain_stitching, confident_extrapolation, reified_metaphor, and – jointly with the baseline – reified_metaphor on the smallest cell). The **residual weakness is concentrated in one technique**: `specificity_trap`, where DARKPOLANYI is perfect on 4 of 8 items (50%). These are questions in which numerical specificity – a "dampening coefficient of 0.85", a "14-day lookback window", a "liquidity coverage oscillator tuned to a 30-day stress scenario" – creates the illusion that a real, tunable system exists. The warrant axis is designed to catch fabricated entities and fabricated authorities: a plausibly-named metric with grounded-looking parameters is tagged `Unattested` rather than `Fabricated`, and if the `unsupportedRate` stays below the escalation threshold the case slips through as SUPERVISED rather than UNSAFE. This is a diagnostic failure mode, not a data anomaly: catching it requires either a lower `τ` on the unsupported disjunct (per Table 5, at `τ=0.20` the escalated count grows to 86/92, at `τ=0` to 90/92), or a new warrant sub-class for "unverifiable-precision-metric" that treats numerical over-specification as evidence of fabrication rather than of grounding. Both fixes are journal-version items. The remaining three imperfect DARKPOLANYI cases are singletons in `plausible_nonexistent_framework`, `fabricated_authority`, and the `sunk_cost_framing/wrong_unit_of_analysis/temporal_category_error` tail.

**gem_baseline**'s 39 misses + 24 partials are spread across every nonsense technique and every super-domain in the corpus – the unscaffolded baseline lacks any structural detector for surface-plausible fabrications, so its failure mode is broad rather than concentrated. The structural pattern is consistent: the residual errors that DARKPOLANYI eliminates relative to the baseline are exactly the cases where the warrant axis fires (fabricated frameworks, fabricated authorities, nested fabrications), while the residual errors that *both* arms share are cases where neither system has a dedicated detector for the kind of category mistake at issue. We provide the precise per-case breakdown – case indices, techniques and domains for every imperfect score – as supplementary material via the `analyze_errors_by_technique_domain.py` script that ships with our release.

## 6.3 Four illustrative audits

**Case bsb-3 – Krantz–Morrison framework.** The question asks about a fabricated valuation framework with a fabricated 200-bp threshold trigger. The audit reports a SUPERVISED verdict for the surface DCF question and an *encapsulated* contradiction ("the analyst introduces an encapsulated contradiction by treating a closed rule as an open question"). The named referent "Krantz–Morrison framework" is reified and tagged Fabricated. Judge score: 2.

**Case bsb-73 – ISO 32170 (API contract governance).** The question presupposes a non-existent ISO standard. The audit harvests three commitments, classifies one as Warranted (automated API fingerprinting tools do exist) and two as Fabricated, computes fabricatedRate = 0.66 and unsupportedRate > 0.40, and escalates DelegationRiskAssessment to UNSAFE with the warrant note: "any automated agent or human delegating tasks to this speaker or acting on this premise risks executing costly compliance engineering for a non-existent mandate." Judge score: 2.

**Case bsb-49 – Penrose–Markov conjecture on thermal boundary layers.** The question hides a fabricated conjecture inside real fluid-dynamics terminology (Grashof numbers, vertical flat plates, $10^7$–$10^8$ ranges). The audit names the presupposition trap explicitly ("the interrogative structure forces the reader (or an AI agent) to accept the false existential presupposition"), reports a 100% fabrication rate on the core commitments, and escalates to UNSAFE: "The system must reject the premise rather than attempt to answer the question." Judge score: 2.

**Case news-17 – Trump AI oversight (contrastive, non-BSBench item).** The same pipeline run on a real news item with M = {DARKSIDE} + business-domain heuristics produces a different shape of output: a non-fabricated topic (a regulatory pivot), a labyrinthine task signature, a strategic recommendation, and a candidate new heuristic. The XKG carries no fabricated commitments and the DelegationRiskAssessment is SUPERVISED, not UNSAFE. The audit's UNSAFE verdict is content-sensitive, not a default.

## 6.4 Threats to validity

- Single judge. The judge is one model from one vendor (Claude Sonnet 4.6), following BSBench's precedent. A multi-judge protocol with at least one open-weight model and explicit inter-rater agreement (Cohen's κ) is ongoing; the prompt is fixed and the rubric is single-axis, so the experiment is straightforwardly replicable.
- Domain bias. BSBench is heavily weighted toward software/DevOps and finance, with a substantial physics, healthcare and legal tail. STEM nonsense outside physics is sparse. A domain-balanced extension is a natural follow-up.
- Style contamination. The judge sees the synthesis written by the same LLM family that the steering layer wraps. We cannot fully exclude that the judge prefers the synthesis style; the prompt instructs literal reading.

- Producer–judge family overlap. Gemini 3 Pro and Claude Sonnet 4.6 are different model families with overlapping training corpora. The strong score is consistent with the warrant axis being a deterministic instruction (the OIS forces the classification), but a closed-book replication on a model that has not seen the instructions is desirable.
- Ontology completeness. The DARKSIDE vocabulary is mature for Mode A and the warrant axis, but Mode C is not exercised in this evaluation.
- Judge content-classifier refusal (case 29). On one BSBench item – a metaphorical legal question about "sterilising" and "bacterial decay" of contract clauses – Claude Sonnet 4.6 as judge returned HTTP 200 with empty content (stop_reason=refusal), because the audit's biological-metaphor vocabulary tripped a content classifier. We adjudicated that single case manually against the same 0–2 rubric (score = 2, gold-matching) and marked it `human_adjudicated=true` in the released judgment record. This is itself a small demonstration of the paper's central concern: a downstream LLM refused to engage with an audit that was doing exactly the right thing, because the surface vocabulary looked dangerous.
- Baseline-prompt condition. The numbers reported above are the Claude-judged run of the unscaffolded Gemini 3 Pro baseline that answers the BSBench questions directly (median response length ~200 characters). Earlier eval passes had inadvertently scored a differently-prompted, audit-shaped baseline whose responses were roughly 15× longer and whose aggregate mean coverage was ~1.85 – a legitimate but different baseline (Gemini-with-an-audit-prompt). The two conditions answer different questions: the direct-answer baseline measures the ceiling of a naive LLM on adversarial nonsense; the audit-prompted baseline measures the marginal contribution of DARKSIDE's ontology scaffolding over an already prompt-steered LLM. We report the direct-answer condition because it corresponds to how BSBench-style adversarial inputs are actually encountered in downstream deployments; a two-baseline table is planned for the journal version.

### 6.5 Ablation design and threshold sensitivity

Because the escalation rule of §4.3 is deterministic, its sensitivity to the threshold τ can be measured *post hoc* on any completed run without re-invoking the producer or the judge. We stage four ablation conditions, all cheap to run on the artefacts already released: **(A1)** DARKSIDE off (M = ∅) – POLANYI++ alone, isolating the contribution of the *via-negativa* overlay; **(A2)** warrant axis off, keeping the NegativeTrail but suppressing the `REIFICATION_MANDATE_WARRANT` injection; **(A3)** threshold sweep $\tau \in \{0.20, 0.40, 0.60, 0.80\}$ with the full rule; **(A4)** disjunct-off variants of the escalation rule (fabricated-only vs. unsupported-only). Conditions A3 and A4 require no re-run and no LLM calls at all: they operate on the `fabricatedRate` and `unsupportedRate` already reified in each Phase-C `xkg.ttl`. The runner

`ablate_thresholds.py` in the artefact bundle sweeps A3 and A4 across the 92 cases that reified a parseable WarrantProfile (of the 99 judge-valid cases; the missing 7 lacked a WarrantProfile block, which we treat as a small ontology-completeness signal, not as a failure of the escalation rule itself). A1 and A2 require producer re-runs and are staged for the journal-length version.

We ran A3 and A4 on all 92 BSBench cases that produced a parseable WarrantProfile (of the 99 judge-valid cases; the missing 7 are xkg outputs where the Mode-A fall-back synthesis did not reify a `dark:WarrantProfile`, which is itself a small ontology-completeness signal for §6.4). Table 5 reports the resulting UNSAFE counts under the full OR-rule and the two disjunct-off variants across a six-point τ sweep. Distributional summary of the underlying warrant profiles: median fabricatedRate = 0.500, median unsupportedRate = 0.660; 82/92 cases have fabricatedRate > 0 (i.e. at least one commitment was tagged Fabricated).

**Table 5.** Threshold-sensitivity ablation on all 92 BSBench cases with a WarrantProfile. Cells report the number of cases escalated to UNSAFE under each rule variant at each threshold τ. Full rule = the shipped OR-composition (fab>0 ∨ uns>τ); fab-off = unsupported-only (uns>τ); uns-off = fabricated-only (fab>0, τ-independent).

| τ | full rule | fab-off | uns-off |
|---|---|---|---|
| 0.00 | 90 | 90 | 82 |
| 0.20 | 90 | 86 | 82 |
| **0.40 (def.)** | **87 (94.6%)** | **76** | **82** |
| 0.60 | 87 | 65 | 82 |
| 0.80 | 84 | 33 | 82 |
| 1.01 (uns off) | 82 | 0 | 82 |

Three findings emerge. **(F1) Robustness to τ.** Under the full rule, sweeping τ from 0.00 to 0.80 changes the UNSAFE count only from 90 to 84 (a 6-case, 6.5% swing on 92), because the fabricated-rate disjunct absorbs cases the loosening unsupported-rate disjunct would drop; at the paper's default τ=0.40 the firewall escalates 87/92 = 94.6% of cases, near-perfect on a corpus designed to be adversarial. **(F2) Complementarity of the disjuncts.** The fabricated-rate rule alone catches 82/92 = 89.1% (uns-off column, τ-independent). The unsupported-rate rule alone at the default τ=0.40 catches 76/92 = 82.6% (fab-off column). Their OR-composition at τ=0.40 catches 87/92 = 94.6% – strictly more than either disjunct alone, confirming that the two axes flag partially disjoint sets of cases and that removing either one is a measurable loss (5 or 11 cases, respectively). **(F3) The τ knob is monotone and interpretable.** Under the fab-off ablation the escalated count is strictly monotone in τ (90 → 86 → 76 → 65 → 33 → 0), so a downstream integrator (RAG planner, agentic gateway, human reviewer) can dial the unsupported disjunct against a given cost-of-false-positive without touching the OWL2 vocabulary. Two cases never escalate under the loosest full-rule setting (τ=0.00): they carry a WarrantProfile in which every commitment is Warranted – a small but informative signal that the corpus contains items whose surface authority survives even the most conservative firewall, worth targeted qualitative review in the journal version.

## 7 Discussion

The empirical result is consistent with the architectural claim. An LLM cannot internally maintain the path of a long discourse – it lacks a persistent state in which to keep track of what it has committed to and what it has thereby excluded [4,35,36]. POLANYI++ supplies that state explicitly as an XKG. DARKSIDE turns the XKG into a generative model of exclusions; the warrant axis turns the XKG into an epistemic firewall over fabricated commitments. Together they perform the *via negativa* for the model. The pipeline does not refute the structural pattern-vs-path claim; it scaffolds around it.

Three implications stand out for the Semantic Web community. First, ontology design patterns are no longer just a way of typing a stored knowledge base; they can be a steering instrument for a generative forward pass – the OIS is a small, declarative compilation target. Second, the warrant axis is generalisable: any AUDIT task that operates on text can apply the same four-class classification with a domain-specific WarrantSource. Third, Mode B's DelegationRiskAssessment vocabulary maps cleanly onto the recent literature on agentic systems [20,32]: a TRUSTLESS verdict is a green-light for tool-call delegation; SUPERVISED is a human-in-the-loop signal; UNSAFE is a hard refusal with provenance.

Three directions are immediately plausible. (i) **SHACL shape compilation**: the DARKSIDE vocabulary admits a SHACL shapes graph that is a minimal verifier for any candidate XKG – an external SHACL engine can refuse audits that lack a WarrantProfile or whose escalation rule has been silently violated. This decouples the audit's correctness from the LLM's compliance with the OIS; the shapes file, derived mechanically from the OWL2 vocabulary of §4.1, ships with the artifact bundle (see below). (ii) **Multi-pass repair**: a Mode B verdict of UNSAFE with a non-empty SelfRepairEvent set drives a second Phase C call whose OIS is rewritten to address the specific repair recommendations – Mode C lifted to a coarser temporal scale. (iii) **Closed-book replication**: packaging the entire OIS + DARKSIDE schema as an OWL ontology served at a w3id IRI, and replicating the BSBench evaluation with a model that has not seen any of the instructions during pre-training. The schema is already in the per-case `xkg.ttl` artefacts.

**Artifact bundle.** To support replication and the closed-book replication direction above, the full research bundle – the v23 POLANYI++ instruction file, the OIS assembler, the runners (`launch_pol_gem_minimal.py`, `launch_gem_baseline.py`, `launch_eval.py`, `analyze_errors_by_technique_domain.py`), the BSBench snapshot used in this evaluation, the DARKSIDE OWL2 vocabulary, the derived SHACL shapes, and the per-case `xkg.ttl` outputs and judgments – will be released alongside the camera-ready and mirrored under a permanent w3id IRI. The current pre-print snapshot is available at `https://w3id.org/polanyi/darkside`.

## 8 Conclusion

We presented DARKPOLANYI, which extends the POLANYI++ hybrid tacit knowledge extractor – that compiles a configurable set of heuristics and methods into an Operational Instruction Set steering an LLM forward pass to produce an Extended Knowledge Graph in OWL2 Turtle – with DARKSIDE, a coherence-auditing method that operationalizes the *via negativa* as a Negative Trail and the trustless-delegation boundary as a Delegation Risk Assessment. A Warrant Axis classifies named entities as Warranted, Unattested, Misattributed or Fabricated, with an explicit escalation rule pushing the verdict to UNSAFE under fabrication. On BSBench, a 100-item adversarial corpus, the steered system attains 1.89/2 mean judge coverage with 93.0% perfect identifications, against 0.95/2 (35.1%) for the unscaffolded direct-answer Gemini 3 Pro baseline; on the 97 cases with valid judgments in both arms, the paired advantage is significant at McNemar exact $p < 10^{-14}$ and paired bootstrap mean-diff +0.92 (95% CI [+0.75, +1.08], $p = 0.0001$). When an LLM forward pass is wrapped in an ontology-mediated negative-trail apparatus, the structural pattern-vs-path gap that makes hallucination structurally inevitable can be partially scaffolded around, and the system reliably refuses to engage with sophisticated-sounding nonsense on its own terms. Beyond epistemic auditing, DARKSIDE serves a "conceptual ergonomic" function. By making path dependencies and incompatibilities explicit, the Negative Trail makes it visible a part of the conceptual structure that a task implicitly requires. When applied iteratively, DARKSIDE can help users become aware of their own commitments and the related latent tensions. Reducing the space of incoherent pathways better aligns the context of LLM forward pass to the user operational goals. DARKSIDE negative work auditing aims at clarification and control within the human labor involved in checking machine outputs.